\documentclass[runningheads]{llncs}
\usepackage{graphicx}
\usepackage{multirow}
\usepackage{amsmath}
\usepackage{amssymb}
\usepackage{booktabs}
\usepackage{tabularx}
\usepackage{makecell}
\usepackage{array}

\begin{document}
\titlerunning{ABDD: Dual-Alignment TTA for Blade Defect Detection}
\title{Robust Online Aero-Engine Blade Defect Detection via Dual-Alignment Test-Time Adaptation}
\author{Zhaoyang Wang\inst{1,2}\orcidID{0000-0002-4944-3924} \and
Haiyong Chen\inst{1}\orcidID{0000-0002-5262-4208} \and
Dongying Li\inst{1}\orcidID{0009-0003-4031-8726} \and
Yining Wang\inst{1}\orcidID{0009-0004-4551-2111} \and
Huapeng Wu\inst{2}\orcidID{0000-0001-6759-6038} \and
Xinwei Lv\inst{1}\orcidID{0009-0000-0976-4590} \and
ATIK SHAHARIAR\inst{1}\orcidID{0009-0006-2841-870X}}
\authorrunning{Z. Wang et al.}
\institute{School of Artificial Intelligence, Hebei University of Technology, Tianjin, China\\ \and Lappeenranta-Lahti University of Technology, Lappeenranta, Finland}
\maketitle            
\begin{abstract}
Reliable visual inspection is essential for quality assurance in aero-engine blade manufacturing, where defect appearance may vary across production lines, imaging conditions, blade poses, and surface backgrounds. Such domain shifts cause a mismatch between training and deployment data and degrade the reliability of deep defect detectors in online inspection. This problem is particularly challenging because aero-engine blade images usually contain sparse defects, making pseudo-label-based adaptation vulnerable to noisy or missing predictions. To address this issue, we propose Aero-engine Blade Defect Detector (ABDD), an online adaptive detection framework based on test-time adaptation. ABDD introduces a Dual-Alignment Strategy to jointly adapt global visual style and local defect morphology by combining feature-statistics alignment with pseudo-box alignment. To reduce error accumulation from unreliable pseudo labels, an Uncertainty-aware Box Filtering mechanism evaluates pseudo boxes using classification confidence, classification entropy, and localization entropy. In addition, a lightweight Sparse Dilated Mona module enables parameter-efficient delta tuning while limiting source-domain forgetting. ABDD is evaluated on CD-AeBD and HD-AeBD under multiple domain-shift scenarios, with TTA strategies compared under a unified RT-DETR + Swin-T architecture. Experiments show that ABDD consistently improves detection robustness under domain shifts, and its practicality is further validated on an industrial inspection platform.

\end{abstract}

\section{Introduction}\label{sec1}
Aero-engine blades (AeBs) are critical components that directly affect engine efficiency, reliability, and flight safety. During manufacturing, subtle surface defects such as scratches, dents, and coating imperfections may occur and later propagate under high mechanical loads, thermal stress, and corrosive environments, leading to structural degradation or even catastrophic failures. Therefore, accurate AeB defect inspection is an essential quality assurance procedure before assembly and deployment \cite{wch,yang2022review}.

\begin{figure}
\centering
\includegraphics[width=0.99\textwidth]{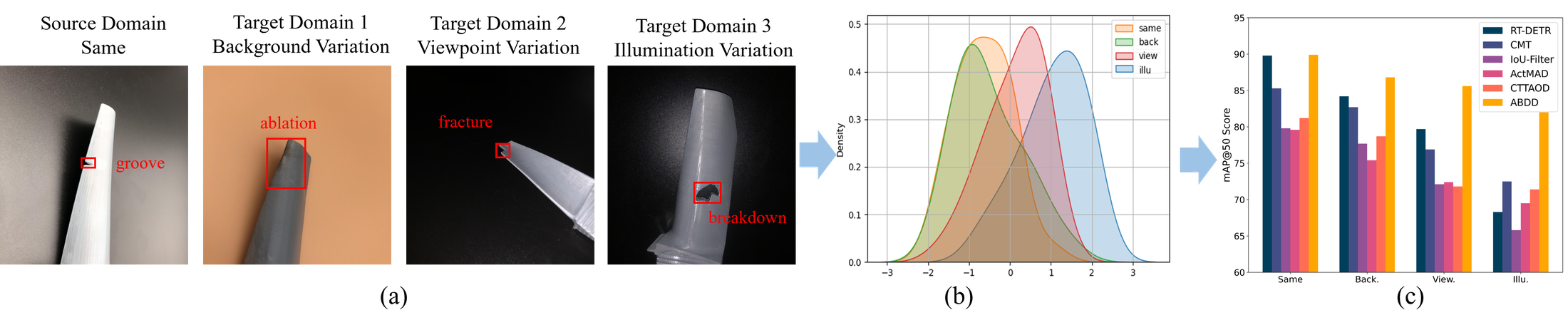}
\caption{Aero-engine blade defect detection under cross-domain shifts. Source and target images may differ in background, viewpoint, and illumination, resulting in feature-distribution discrepancies and detection performance degradation.}
\label{fig_1}
\end{figure}

Conventional AeB inspection mainly relies on manual visual examination or rule-based machine vision, which are labor-intensive, subjective, and sensitive to illumination changes, surface textures, and complex defect morphologies. Recently, deep learning has provided an effective solution for automated industrial defect inspection \cite{wang2020single,ShangHB_TII}, enabling different AeB defect types to be localized and classified simultaneously through object detection. However, deployed detectors often suffer from domain shifts at inference time, where target-domain data differ from the source-domain training distribution and violate the independent and identically distributed assumption \cite{zhang2020towards,cpdd}. As shown in Fig. 1, subtle changes in background, viewpoint, or illumination can still lead to clear feature-distribution discrepancies and performance degradation \cite{AeBAD}. Although domain adaptation can alleviate such gaps, it usually requires offline retraining with sufficient target-domain data, which is impractical for AeB production because working conditions change frequently and defect samples are sporadic \cite{oza2023unsupervised}. Therefore, we adopt test-time adaptation (TTA), which updates a pre-trained detector online using unlabeled test samples from unknown target domains \cite{liang2025comprehensive}. Existing detection-oriented TTA methods mainly rely on pseudo-box learning or feature-statistic alignment \cite{CTTAOD,doan2024weakly,ActMAD}, but they are mostly designed for autonomous driving or common-object style shifts \cite{stfar,test,AMROD}, and are not directly suitable for fine-grained AeB defect detection.

% \begin{figure}[t]
% \centering
% \includegraphics[width=1.0\textwidth]{Images/Fig2.png}
% \vspace{-0.1cm}
% \caption{Comparison of different TTA strategies. Purple arrows denote gradient backpropagation, and orange modules indicate the components introduced in our method.}
% \label{fig_2}
% \vspace{-0.1cm}
% \end{figure}

Existing TTA methods face two main challenges in AeB defect detection. First, pseudo-box-based methods rely on abundant and reliable pseudo labels \cite{wangshijie}, whereas AeB images usually contain fewer than one defect per image, making normal images uninformative and erroneous pseudo boxes harmful. Second, many feature-alignment methods are designed for Faster R-CNN-style detectors with ROI pooling \cite{faster}, limiting their compatibility with modern YOLO \cite{yolov10,yolov13} and DETR-based detectors \cite{rtdetr,DEIM}. Transformer backbones \cite{Swin} further increase adaptation difficulty due to their large parameter scale and weak inductive bias.

To address these issues, we hypothesize that pseudo boxes and feature statistics can complementarily adapt local and global domain shifts. Local shifts mainly arise from defect morphology variations caused by blade type or viewpoint changes, where pseudo-label supervision is useful, while global shifts caused by background or illumination can be mitigated by feature-statistics alignment using both normal and defective target images. Based on this hypothesis, we propose the \textbf{Aero-engine Blade Defect Detector (ABDD)}, an online adaptive detector with a \textbf{Dual-Alignment Strategy (DAS)} in a teacher--student framework. During test time, the teacher generates pseudo boxes and computes feature-alignment losses against source-domain statistics, while the student is updated by pseudo-box and feature-alignment objectives. The teacher is updated by exponential moving average. To improve pseudo-label quality, \textbf{Uncertainty-aware Box Filtering (UBF)} removes predictions with high classification or localization uncertainty. To reduce adaptation cost and avoid catastrophic forgetting, \textbf{Sparse Dilated Mona (SDM)} performs delta tuning by freezing most parameters and updating only lightweight modules.

We evaluate ABDD on two self-collected datasets, CD-AeBD and HD-AeBD. On CD-AeBD Subset I, ABDD improves $mAP@50$ by 13.7\%, 5.9\%, and 2.6\% under illumination variation, viewpoint change, and background shift, respectively. On CD-AeBD Subset II, where viewpoint and illumination change simultaneously, ABDD achieves an average 11.5\% $mAP@50$ gain. On HD-AeBD, ABDD improves $mAP@50$ and $mAP@50:95$ by 2.8\% and 6.4\%, respectively. ABDD is further integrated into a robotic-arm-based inspection system deployed with a major aero-engine manufacturer. The main contributions are summarized as follows:

\begin{itemize}
    \item We identify the domain-shift problem in AeB defect detection and analyze why existing TTA methods are insufficient for this industrial scenario.
    \item We propose ABDD, which combines DAS, UBF, and SDM to jointly exploit pseudo-box supervision, feature-statistics alignment, and efficient delta tuning during test-time adaptation.
    \item We curate two AeB defect datasets, CD-AeBD and HD-AeBD, and conduct extensive experiments to validate the effectiveness and robustness of ABDD.
\end{itemize}

\section{Related work}
\subsection{Defect Detection of Aero-engine Blades}
Early AeB defect detection mainly relied on manual inspection, traditional machine learning, edge/contour analysis, and ultrasonic imaging or echo features \cite{aust2021automated,9337557,jiao2021novel}, but these methods are usually limited in robustness and generalization. Recent studies have introduced CNN- and Transformer-based detectors for AeB inspection. For example, improved Faster R-CNN, DETR-based lightweight detectors, cross-layer semantic guidance, superpixel graph modeling, and coarse-to-fine localization have been explored to detect minute, discontinuous, or irregular defects \cite{yixuan2023aeroengine,sun2024sdd,shang2024superpixel,wch}. However, most of them assume relatively fixed data distributions and mainly focus on subtle-defect perception or model efficiency. Although Zhang et al. \cite{AeBAD} investigate multi-domain AeB anomaly detection, their method is trained without defective samples and does not address online adaptation for defect detection under changing target domains.

\subsection{Test-Time Adaptation for Object Detectors}
Test-time adaptation (TTA) updates a pre-trained detector during inference to handle changing target-domain distributions. Existing detection-oriented TTA methods mainly follow two lines: feature alignment and pseudo-box learning. Feature-alignment methods reduce domain gaps by matching target-domain statistics with source or historical statistics, such as adapting BN or RoI-related parameters \cite{test,tent}, aligning feature means and variances \cite{ActMAD}, or minimizing feature-batch divergence with lightweight adapters \cite{CTTAOD}. Pseudo-box-based methods usually adopt a teacher--student framework, where the teacher generates pseudo labels and the student is updated with filtered predictions \cite{dobler2023robust}. Representative works improve pseudo-label usage through memory prototypes and contrastive learning \cite{vs2023towards}, IoU- and overlap-based filtering \cite{IoUFilter}, or human-in-the-loop weak supervision \cite{doan2024weakly}. However, these methods are mostly developed for autonomous driving or common-object scenarios with dense pseudo labels. AeB defect detection instead contains sparse defects, dominant normal samples, and coupled domain shifts, motivating ABDD to jointly exploit pseudo-box supervision, feature-statistics alignment, and lightweight delta tuning.

\section{Methodology}
\subsection{Aero-engine Blade Defect Detector}
As shown in Fig. 2, ABDD follows a DETR-style architecture with a backbone, Transformer encoder, Transformer decoder, and detection head. The backbone consists of four cascaded Swin with Sparse Dilated Mona (SSDM) modules, where SDM enables parameter-efficient adaptation on top of Swin Transformer~\cite{Swin}. The encoder and decoder follow RT-DETR~\cite{rtdetr} for efficient multiscale feature fusion and deformable cross-attention between image features and object queries. The detection head contains separate classification and localization branches. The localization branch adopts FCOS-style regression, predicting distances $(l,t,r,b)$ from sampling points to box edges~\cite{fcos}, which facilitates uncertainty-aware box filtering.

\begin{figure}
\centering
\includegraphics[width=\textwidth]{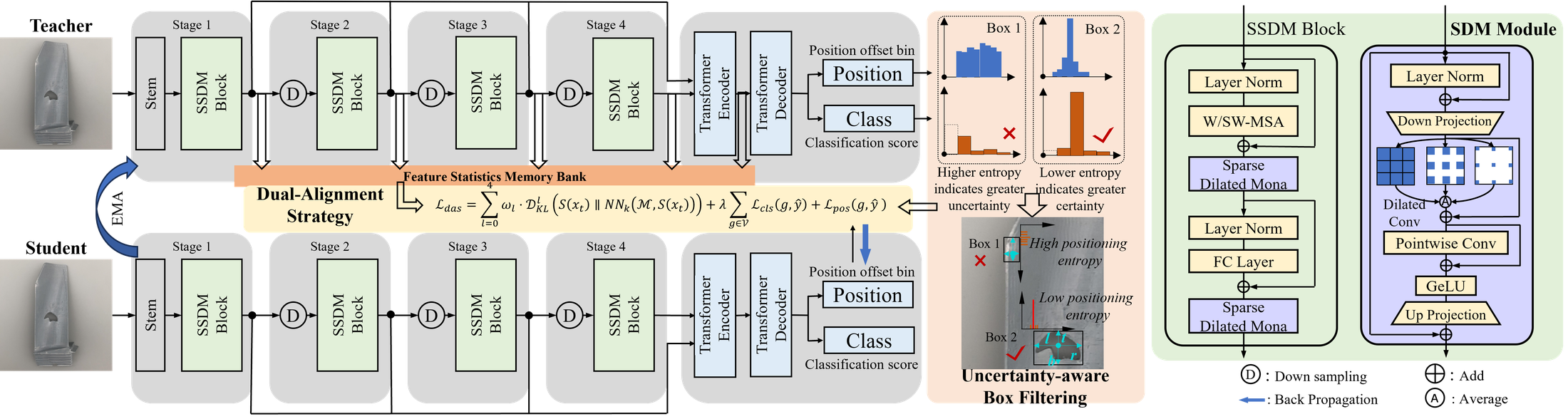}
\caption{Overview of the proposed ABDD for aero-engine blade defect detection.}
\label{fig_3}
\end{figure}

\subsection{Dual-Alignment Strategy}
Given a source-trained detector $f_{\theta^*}$ trained on labeled source data $D_{\text{train}}=\{(x_i,y_i)\}_{i=1}^{N}$, where $y_i=(b_i,c_i)$, test-time adaptation aims to update the model on unlabeled target samples $D_{\text{test}}=\{x_j\}_{j=1}^{M}$ with $x_j\sim P_{\text{test}}(x)$ and $P_{\text{test}}(x)\neq P_{\text{train}}(x)$. Starting from $\theta_0=\theta^*$, the online adaptation process is formulated as
\begin{equation}
\theta_t = \theta_{t-1} - \eta \nabla_{\theta} 
\mathcal{L}_{\text{adapt}}\big(f_{\theta_{t-1}}(x_t)\big),
\end{equation}
where $x_t$ is the current target sample, $\eta$ is the learning rate, and $\mathcal{L}_{\text{adapt}}$ is an unsupervised adaptation loss. In ABDD, $\mathcal{L}_{\text{adapt}}$ is built with a Dual-Alignment Strategy (DAS), which jointly aligns pseudo boxes and intermediate feature statistics to handle local defect-level and global image-level domain shifts.

\subsubsection{Pseudo-Box Alignment}

To adapt to defect morphology changes, we use a teacher--student framework for pseudo-box alignment. The teacher generates pseudo boxes $\{(b_i,c_i)\}$ on target images, and UBF removes uncertain predictions before they supervise the student. The pseudo-box alignment loss is defined as
\begin{equation}
\mathcal{L}_{\text{pse}} =
\sum_{i=1}^{N_p} \mathrm{UBF}(b_i,c_i)
\left[
\mathcal{L}_{\text{cls}}(c_i,\hat{c}_i)
+
\mathcal{L}_{\text{reg}}(b_i,\hat{b}_i)
\right],
\end{equation}
where $N_p$ is the number of pseudo boxes, $\mathrm{UBF}(b_i,c_i)\in\{0,1\}$ indicates whether the pseudo box is retained, and $\hat{c}_i$ and $\hat{b}_i$ are the student predictions. The classification loss adopts focal loss, and the regression loss combines GIoU and L1 losses:
\begin{equation}
\mathcal{L}_{\text{reg}} =
\lambda_{\text{giou}}\left(1-\mathrm{GIoU}(b_i,\hat{b}_i)\right)
+
\lambda_{\text{l1}}\lVert b_i-\hat{b}_i\rVert_1 .
\end{equation}
Pseudo-box alignment transfers source-domain detection knowledge to the target domain at the instance level.

\subsubsection{Feature Distribution Alignment}
To handle global domain shifts such as illumination and background variations, we introduce feature distribution alignment. A source-domain memory bank $\mathcal{M}$ is constructed offline after source training by forwarding source images through the frozen detector and storing multi-level channel-wise feature statistics:
\begin{equation}
m_i = \{(\mu_i^l,\sigma_i^l)\}_{l=1}^{L}.
\end{equation}
To reduce storage and retrieval cost, at most $\mathcal{B}$ statistic vectors are retained by class-balanced random sampling, with $\mathcal{B}=500$ by default.

During test-time adaptation, the teacher extracts the statistic vector of the current target image $x_t$ as the query, retrieves its $k$ nearest source statistics from $\mathcal{M}$ according to Euclidean distance, and aligns the target statistics with the averaged retrieved statistics. The feature alignment loss is defined as
\begin{equation}
\mathcal{L}_{\mathrm{fea}}
=
\sum_{l=1}^{L}
\omega_l
\mathcal{D}_{\mathrm{KL}}^{l}.
\end{equation}
Here, $\omega_l$ is the weight of the $l$-th feature level, and $\mathcal{D}_{\mathrm{KL}}^{l}$ denotes the KL divergence between the target feature distribution and the retrieved source feature distribution at level $l$:
\begin{equation}
\mathcal{D}_{\mathrm{KL}}^{l}
=
D_{\mathrm{KL}}
\left(
\mathcal{N}(\mu_t^l,(\sigma_t^l)^2),
\mathcal{N}(\mu_m^l,(\sigma_m^l)^2)
\right).
\end{equation}
where $(\mu_t^l,\sigma_t^l)$ are the target-domain feature statistics, and $(\mu_m^l,\sigma_m^l)$ are the averaged statistics of the retrieved $k$ nearest source samples, with $k=2$ by default. Unlike pseudo-box alignment, feature distribution alignment can also exploit normal target images and thus complements instance-level adaptation. The overall DAS objective is
\begin{equation}
\mathcal{L}_{\mathrm{das}}
=
\mathcal{L}_{\mathrm{pse}}
+
\mathcal{L}_{\mathrm{fea}}.
\end{equation}
To improve efficiency and reduce catastrophic forgetting, only the SDM modules and detection head of the student model are updated, while the remaining parameters are frozen:
\begin{equation}
\phi_{\mathrm{student}}
\leftarrow
\phi_{\mathrm{student}}
-
\eta
\nabla_{\phi_{\mathrm{student}}}
\mathcal{L}_{\mathrm{das}}.
\end{equation}
The teacher is updated by exponential moving average of the student parameters:
\begin{equation}
\phi_{\mathrm{teacher}}
\leftarrow
\alpha\phi_{\mathrm{teacher}}
+
(1-\alpha)\phi_{\mathrm{student}}.
\end{equation}

\subsection{Uncertainty-aware Box Filtering}
To reduce error accumulation from unreliable pseudo boxes, we introduce Uncertainty aware Box Filtering (UBF), which jointly considers classification confidence, classification entropy, and localization entropy. For each object query, the detection head predicts classification probabilities, center coordinates $(x,y)$, and offset-bin distributions for four boundaries $(l,t,r,b)$. The localization and classification entropies are computed as
\begin{equation}
H_{\mathrm{loc}}
=
-\frac{1}{4}
\sum_{k \in \{l,t,r,b\}}
\sum_{j=1}^{K}
q_{j}^{k}\log q_{j}^{k},
\end{equation}
\begin{equation}
H_{\mathrm{cls}}
=
-\sum_{c=1}^{C}
p_c\log p_c,
\end{equation}
where $q_j^k$ denotes the probability of the $j$-th offset bin for boundary $k$, and $p_c$ is the predicted probability of category $c$. In our implementation, $K=16$ and $C$ is dataset-dependent, defined as the number of defect categories plus one background class. The entropies are normalized as
\begin{equation}
\widetilde{H}_{\mathrm{loc}}
=
\frac{H_{\mathrm{loc}}}{\log K},
\qquad
\widetilde{H}_{\mathrm{cls}}
=
\frac{H_{\mathrm{cls}}}{\log C}.
\end{equation}
A pseudo box is retained only when
\begin{equation}
\mathrm{UBF}(b,c)
=
\mathbb{I}
\left[
\widetilde{H}_{\mathrm{loc}} < 0.5,
\;
\widetilde{H}_{\mathrm{cls}} < 0.5,
\;
\mathrm{conf} > 0.25
\right],
\end{equation}
where $\mathrm{conf}=\max_c p_c$. Thus, UBF suppresses boxes with uncertain categories or ambiguous localization, improving the reliability of test-time supervision.

\subsection{Sparse Dilated Mona}

Although DETR-style detectors avoid post-processing such as non-maximum suppression, their Transformer backbones are expensive to adapt during test time. To enable efficient and stable adaptation, we freeze the backbone and update only the proposed Sparse Dilated Mona (SDM) modules. As shown in Fig. 2, each SDM contains layer normalization, downsampling projection, parallel dilated convolutions, pointwise convolution, GELU activation, upsampling projection, and a residual connection:
\begin{equation}
x^{l+1}
=
x^{l}
+
U
\left(
\sigma
\left(
F_{\mathrm{pw}}
\left(
F_{\mathrm{dc}}
\left(
D
\left(
\mathrm{LN}(x^{l})
\right)
\right)
\right)
\right)
\right),
\end{equation}
where $D$ reduces the channel dimension from $d$ to $n=d/4$, $U$ restores it to $d$, $F_{\mathrm{dc}}$ denotes parallel depthwise separable dilated convolutions, $F_{\mathrm{pw}}$ denotes pointwise convolution, and $\sigma$ is GELU.

Inspired by Mona~\cite{Mona}, two SDM modules are inserted into each Swin block to form the SSDM backbone. Each SDM uses three parallel branches: a standard $3 \times 3$ convolution, a $3 \times 3$ dilated convolution with dilation rate 2, and a $3 \times 3$ dilated convolution with dilation rate 3, corresponding to effective receptive fields of $3 \times 3$, $5 \times 5$, and $7 \times 7$, respectively. Compared with explicit $3 \times 3$, $5 \times 5$, and $7 \times 7$ convolutions, SDM reduces the trainable parameters of the parallel convolution branches from $2(3^2+5^2+7^2)n=166n$ to $2 \times 3 \times 3^2 n=54n$, saving $112n$ parameters. Therefore, SDM lowers adaptation cost while preserving source-domain knowledge and mitigating catastrophic forgetting. Additional method details, including the comparison with existing TTA paradigms and the pseudo-code of ABDD, are provided in \textit{Supplementary Material~III}.

\section{Experimental evaluation}
\subsection{Datasets and Experimental Settings}

To evaluate ABDD under continuously changing domain shifts, we use three evaluation settings from two aero-engine blade defect datasets: CD-AeBD I, CD-AeBD II, and HD-AeBD. CD-AeBD I is derived from AeBAD-S~\cite{AeBAD}, where pixel-level anomaly masks are converted into bounding boxes by connected-component extraction. CD-AeBD II is derived from AeBAD-V and is manually annotated with bounding boxes. HD-AeBD is collected from real industrial inspection scenes using high-definition cameras and contains blade images from different production batches. Detailed construction protocols, including mask-to-box conversion, annotation rules, splits, category statistics, domain definitions, and data availability, are provided in \textit{supplementary material I}. All datasets are organized in the VOC format, and their statistics are summarized in Table 1.

\begin{table}[t]
\centering
\caption{Summary of the datasets and domain-shift settings. Def. denotes defect categories.}
\label{tab_dataset_summary}
\scriptsize
\setlength{\tabcolsep}{3pt}
\renewcommand{\arraystretch}{1.15}
\begin{tabularx}{\textwidth}{@{}lccclX@{}}
\toprule
Setting & Images & Boxes & Def. & Domains & Source $\rightarrow$ Target \\
\midrule
CD-AeBD I  
& 1,639 
& 2,371 
& 4 
& \makecell[l]{Same, Back.,\\View., Illum.} 
& Same $\rightarrow$ Back./View./Illum. \\

CD-AeBD II 
& 2,703 
& 4,050 
& 4 
& D1, D2, D3 
& D1 $\rightarrow$ D2/D3 \\

HD-AeBD    
& 1,857 
& 2398 
& 3 
& B1, B2, B3 
& B1 $\rightarrow$ B2/3 \\
\bottomrule
\end{tabularx}
\vspace{-0.2cm}
\end{table}

All experiments are implemented with Python 3.10 and PyTorch 2.1 on two NVIDIA RTX 4090 GPUs. ABDD uses RT-DETR with a Swin-T backbone as the baseline detector. The images are resized to $640 \times 640$, and no data augmentation is used. The source pretraining adopts AdamW with a learning rate of $1\times10^{-4}$ and batch size 8, while TTA uses AdamW with a learning rate of $1\times10^{-5}$. Following the TTA protocol, target domain images are processed sequentially in a streaming manner, and the model continuously performs inference and adaptation rather than adapting to each target domain separately.

\subsection{Main Results}
\subsubsection{Quantitative Results}
We compare seven source-only object detectors and seven TTA strategies, with results reported in Table 2 and Table 3. For a fair comparison, all TTA strategies are implemented and evaluated under the same detector architecture as ABDD, i.e., RT-DETR with a Swin-T backbone. Therefore, the reported results reflect the effectiveness of different adaptation strategies rather than differences in detector backbones. Specifically, \textbf{CMT} denotes contrastive mean-teacher pseudo-box alignment; \textbf{IoU-Filter} uses multi-round detection consistency for pseudo-label filtering; \textbf{ActMAD} performs feature-distribution alignment by minimizing source-target activation discrepancy; \textbf{CTTAOD} adopts adapter-based feature alignment for continual test-time adaptation; \textbf{AMROD} combines mean-teacher pseudo-labeling with adaptive monitoring and randomized restoration; \textbf{STFAR} integrates self-training with feature-alignment regularization; and \textbf{SMDC} introduces semantic mining and domain correction into teacher-student pseudo-box adaptation.

\begin{table}
  \centering
  \caption{Results ($mAP@50$) of detectors and TTA methods across different domains in CD-AeBD I. \textit{Same}, \textit{Back.}, \textit{View.}, and \textit{Illum.} denote source, background-shift, viewpoint-shift, and illumination-shift domains, while \textit{Ab}, \textit{Br}, \textit{Fr}, and \textit{Gr} denote ablation, breakdown, fracture, and groove. The \textit{mean} denotes the average score across four domains, representing overall performance. Best scores are highlighted in \textbf{bold}.}
  \label{tab_3}
  \resizebox{\textwidth}{!}{%
    \begin{tabular}{p{6.8em}|cccc|cccc|ccc}
    \toprule
    Method & Same & Back. & View. & Illum. & Ab & Br & Fr & Gr & Mean & F1 & FPS \\
    \midrule
    Yolov10~\cite{yolov10} 
    & 87.9 & 82.0$_{\downarrow5.9}$ & 77.4$_{\downarrow10.5}$ & 69.7$_{\downarrow18.2}$ 
    & 67.7 & 86.5 & 83.3 & 80.8 & 79.3 & 70.7 & 42.5 \\

    Yolov13~\cite{yolov13} 
    & 86.8 & 80.8$_{\downarrow6.0}$ & 77.9$_{\downarrow8.9}$ & 67.4$_{\downarrow19.4}$ 
    & 78.2 & 85.6 & 72.0 & 78.9 & 78.2 & 69.9 & 43.8 \\

    RT-DETR~\cite{rtdetr}  
    & 89.8 & 84.2$_{\downarrow5.6}$ & 79.7$_{\downarrow10.1}$ & 68.3$_{\downarrow21.5}$  
    & 68.8 & 87.2 & 83.1 & 82.0 & 80.5 & 71.4 & 22.9 \\

    MI-DETR~\cite{MIDETR} 
    & 89.8 & 85.5$_{\downarrow4.3}$ & 79.0$_{\downarrow10.8}$ & 75.5$_{\downarrow14.3}$ 
    & 80.7 & 83.2 & 86.8 & 77.7 & 82.5 & 73.6 & 20.2 \\

    Dfine~\cite{dfine}
    & 90.4 & 86.5$_{\downarrow3.9}$ & 80.9$_{\downarrow9.5}$ & 71.1$_{\downarrow19.3}$ 
    & 73.6 & 87.8 & 84.0 & 83.2 & 82.2 & 72.3 & 36.0 \\

    MR-DETR~\cite{MRDETR}  
    & \textbf{92.1} & \textbf{87.0}$_{\downarrow5.1}$ & 83.2$_{\downarrow8.9}$ & 76.6$_{\downarrow15.5}$ 
    & 77.1 & 87.6 & 85.7 & \textbf{86.9} & 84.7 & 76.7 & 10.8 \\

    DEIM~\cite{DEIM} 
    & 89.5 & 82.1$_{\downarrow7.4}$ & 71.8$_{\downarrow17.7}$ & 70.6$_{\downarrow18.9}$ 
    & 77.4 & 82.3 & 83.2 & 72.7 & 78.5 & 71.0 & 37.5 \\
    \midrule

    CMT~\cite{meanteacher}  
    & 85.3 & 82.7$_{\downarrow2.6}$ & 76.9$_{\downarrow8.4}$ & 72.5$_{\downarrow12.8}$ 
    & 70.1 & 85.6 & 80.8 & 83.4 & 79.4 & 64.4 & 4.2 \\

    IoU-Filter~\cite{IoUFilter}  
    & 79.8 & 77.7$_{\downarrow2.1}$ & 72.1$_{\downarrow7.7}$ & 65.8$_{\downarrow14.0}$ 
    & 75.8 & 72.0 & 70.7 & 76.9 & 73.9 & 60.8 & 1.8 \\

    ActMAD~\cite{ActMAD} 
    & 79.6 & 75.4$_{\downarrow4.2}$ & 72.4$_{\downarrow7.2}$ & 69.5$_{\downarrow10.1}$ 
    & 74.2 & 72.3 & 76.4 & 71.1 & 74.2 & 61.7 & 10.4 \\

    CTTAOD~\cite{CTTAOD} 
    & 81.2 & 78.7$_{\downarrow2.5}$ & 71.8$_{\downarrow9.4}$ & 71.4$_{\downarrow9.8}$  
    & 78.8 & 73.6 & 77.2 & 68.8 & 75.8 & 63.9 & 6.6 \\

    AMROD~\cite{AMROD} 
    & 82.1 & 78.4$_{\downarrow3.7}$ & 73.2$_{\downarrow8.9}$ & 70.1$_{\downarrow12.0}$  
    & 71.9 & 76.5 & 81.8 & 77.0 & 76.1 & 68.5 & 11.2 \\

    STFAR~\cite{stfar}  
    & 84.0 & 80.7$_{\downarrow3.3}$ & 76.2$_{\downarrow7.8}$ & 74.1$_{\downarrow9.9}$  
    & 81.9 & 77.7 & 75.2 & 80.1 & 78.8 & 66.4 & 3.1 \\

    SMDC~\cite{wangshijie}  
    & 80.6 & 77.1$_{\downarrow3.5}$ & 70.4$_{\downarrow10.2}$ & 67.2$_{\downarrow13.4}$  
    & 71.8 & 74.0 & 69.8 & 79.5 & 73.8 & 60.9 & 2.4 \\
    \midrule

    ABDD  
    & 89.9 & 86.8$_{\downarrow3.1}$ & \textbf{85.6}$_{\downarrow4.3}$ & \textbf{82.0}$_{\downarrow7.9}$  
    & \textbf{84.9} & \textbf{89.8} & \textbf{87.5} & 85.2 & \textbf{86.1} & \textbf{81.3} & 5.7 \\
    \bottomrule
    \end{tabular}%
  }
  \vspace{-0.2cm}
\end{table}

As shown in Table 2, non-adaptive detectors generally suffer clear performance drops when transferred from the source domain to target domains with background, viewpoint, or illumination shifts. Existing TTA methods can alleviate this degradation to some extent, but their improvements are not always stable, especially for defect categories with sparse or ambiguous pseudo boxes. In contrast, ABDD achieves more consistent adaptation across target domains by combining pseudo-box supervision with feature-statistics alignment.

In CD-AeBD I, ABDD improves over the baseline detector by 2.6\%, 5.9\%, and 13.7\% on Back., View., and Illum., respectively, and surpasses the best-performing TTA baseline CMT by 6.7\% in mean $mAP@50$. ABDD also achieves at least 84.9\% AP on all four defect categories, with 84.9\%, 89.8\%, 87.5\%, and 85.2\% for Ab, Br, Fr, and Gr, respectively. In comparison, several pseudo-box-based TTA methods show obvious degradation on specific categories, suggesting that pseudo-box supervision alone is vulnerable to noisy predictions under domain shift. These results indicate that DAS and UBF jointly improve both domain robustness and defect-level reliability.

Although ABDD is slower than feed-forward detectors, its online throughput is acceptable for the inspection cycle specified by our industrial partner; details of the deployed validation platform and timing analysis are provided in \textit{Supplementary Material II}.

Table 3 summarizes the results on CD-AeBD II and HD-AeBD. For CD-AeBD II, D1 is used as the source domain, while D2 and D3 are target domains with coupled viewpoint and illumination shifts. Non-adaptive detectors show clear degradation on both target domains, with most performance drops exceeding 16\%. Several existing TTA methods, such as CMT, ActMAD, and AMROD, also perform unstably under such large domain gaps, and their adapted results can even be lower than those of non-adaptive detectors. In contrast, ABDD maintains $mAP@50$ values of 78.3\% and 75.8\% on D2 and D3, respectively, with degradation below 10\%, achieving the best mean $mAP@50$ of 79.9\%. For HD-AeBD, B1 is the source domain and B2/B3 are target domains from different production batches. Under the stricter $mAP@50:95$ metric, ABDD achieves the highest mean score of 44.4\% and the smallest drops on both target batches, indicating stronger robustness for high-precision localization in real industrial inspection scenarios.

\begin{table}
\centering
\caption{Results of state-of-the-art detectors and TTA methods on CD-AeBD II and HD-AeBD. CD-AeBD II reports $mAP@50$ across D1--D3. HD-AeBD reports $mAP@50$ and $mAP@50:95$ across B1--B3. Subscripted downward arrows indicate performance drops from the source domain. The best scores in each column are highlighted in \textbf{bold}.}
\label{tab_cd2_hd_results}
% \tiny
\setlength{\tabcolsep}{2.2pt}
\renewcommand{\arraystretch}{0.88}
\resizebox{\textwidth}{!}{%
\begin{tabular}{l|cccc|cccc|cccc}
\toprule
\multirow{2}{*}{Method}
& \multicolumn{4}{c|}{CD-AeBD II ($mAP@50$)}
& \multicolumn{4}{c|}{HD-AeBD ($mAP@50$)}
& \multicolumn{4}{c}{HD-AeBD ($mAP@50:95$)} \\
\cmidrule(lr){2-5}
\cmidrule(lr){6-9}
\cmidrule(lr){10-13}
& D1 & D2 & D3 & Mean
& B1 & B2 & B3 & Mean
& B1 & B2 & B3 & Mean \\
\midrule

Yolov10~\cite{yolov10}
& 85.2 & 63.1$_{\downarrow22.1}$ & 65.4$_{\downarrow19.8}$ & 71.2
& 88.4 & 75.0$_{\downarrow13.4}$ & 78.8$_{\downarrow9.6}$ & 80.7
& 50.8 & 36.1$_{\downarrow14.7}$ & 35.6$_{\downarrow15.2}$ & 40.8 \\

Yolov13~\cite{yolov13}
& 84.9 & 66.0$_{\downarrow18.9}$ & 68.7$_{\downarrow16.2}$ & 73.2
& 85.7 & 68.9$_{\downarrow16.8}$ & 71.2$_{\downarrow14.5}$ & 75.3
& 45.8 & 29.7$_{\downarrow16.1}$ & 26.8$_{\downarrow19.0}$ & 34.1 \\

RT-DETR~\cite{rtdetr}
& 85.7 & 60.3$_{\downarrow25.4}$ & 59.3$_{\downarrow26.4}$ & 68.4
& 86.3 & 75.2$_{\downarrow11.1}$ & 77.3$_{\downarrow9.0}$ & 79.6
& 47.8 & 32.3$_{\downarrow15.5}$ & 33.9$_{\downarrow13.9}$ & 38.0 \\

Dfine~\cite{dfine}
& 84.8 & 64.9$_{\downarrow19.9}$ & 66.0$_{\downarrow18.8}$ & 71.9
& 83.7 & 74.4$_{\downarrow9.3}$ & 69.9$_{\downarrow13.8}$ & 76.0
& 47.4 & 34.0$_{\downarrow13.4}$ & 31.4$_{\downarrow16.0}$ & 37.6 \\

MR-DETR~\cite{MRDETR}
& \textbf{87.9} & 68.8$_{\downarrow19.1}$ & 70.8$_{\downarrow17.1}$ & 75.8
& \textbf{89.0} & 80.8$_{\downarrow8.2}$ & 79.2$_{\downarrow9.8}$ & \textbf{83.0}
& \textbf{56.1} & 29.6$_{\downarrow26.5}$ & 30.8$_{\downarrow25.3}$ & 38.8 \\

MI-DETR~\cite{MIDETR}
& 87.1 & 70.0$_{\downarrow17.1}$ & 69.4$_{\downarrow17.7}$ & 75.5
& 88.8 & 77.3$_{\downarrow11.5}$ & 76.8$_{\downarrow12.0}$ & 81.0
& 55.9 & 32.5$_{\downarrow23.4}$ & 38.7$_{\downarrow17.2}$ & 42.4 \\

DEIM~\cite{DEIM}
& 83.3 & 65.3$_{\downarrow18.0}$ & 67.0$_{\downarrow16.3}$ & 71.9
& 81.7 & 71.1$_{\downarrow10.6}$ & 70.8$_{\downarrow10.9}$ & 74.5
& 43.7 & 28.0$_{\downarrow15.7}$ & 26.6$_{\downarrow17.1}$ & 32.8 \\
\midrule

CMT~\cite{meanteacher}
& 84.8 & 48.1$_{\downarrow36.7}$ & 24.5$_{\downarrow60.3}$ & 52.5
& 85.3 & 78.4$_{\downarrow6.9}$ & 69.7$_{\downarrow15.6}$ & 77.8
& 46.6 & 40.5$_{\downarrow6.1}$ & 32.1$_{\downarrow14.5}$ & 39.7 \\

IoU-Filter~\cite{IoUFilter}
& 77.2 & 59.5$_{\downarrow17.7}$ & 44.7$_{\downarrow32.5}$ & 60.5
& 79.7 & 73.0$_{\downarrow6.7}$ & 70.2$_{\downarrow9.5}$ & 74.3
& 36.7 & 31.5$_{\downarrow5.2}$ & 29.4$_{\downarrow7.3}$ & 32.5 \\

ActMAD~\cite{ActMAD}
& 70.9 & 45.6$_{\downarrow25.3}$ & 33.8$_{\downarrow37.1}$ & 50.1
& 77.2 & 68.6$_{\downarrow8.6}$ & 53.8$_{\downarrow23.4}$ & 66.5
& 33.2 & 27.8$_{\downarrow5.4}$ & 24.9$_{\downarrow8.3}$ & 28.6 \\

CTTAOD~\cite{CTTAOD}
& 71.6 & 52.6$_{\downarrow19.0}$ & 50.1$_{\downarrow21.5}$ & 58.1
& 78.7 & 71.7$_{\downarrow7.0}$ & 67.4$_{\downarrow11.1}$ & 72.6
& 40.3 & 33.2$_{\downarrow7.1}$ & 30.4$_{\downarrow9.9}$ & 34.6 \\

AMROD~\cite{AMROD}
& 72.3 & 50.1$_{\downarrow22.2}$ & 28.6$_{\downarrow43.7}$ & 50.3
& 76.3 & 68.5$_{\downarrow7.8}$ & 61.9$_{\downarrow14.4}$ & 68.9
& 37.9 & 31.2$_{\downarrow6.7}$ & 27.4$_{\downarrow10.5}$ & 32.2 \\

STFAR~\cite{stfar}
& 73.0 & 60.7$_{\downarrow12.3}$ & 51.4$_{\downarrow21.6}$ & 61.7
& 80.4 & 74.3$_{\downarrow6.1}$ & 69.6$_{\downarrow10.8}$ & 74.8
& 42.2 & 36.4$_{\downarrow5.8}$ & 33.6$_{\downarrow8.6}$ & 37.4 \\

SMDC~\cite{wangshijie}
& 71.6 & 52.8$_{\downarrow18.8}$ & 41.0$_{\downarrow30.6}$ & 55.1
& 78.5 & 69.3$_{\downarrow9.2}$ & 65.2$_{\downarrow13.3}$ & 71.0
& 40.5 & 33.2$_{\downarrow9.1}$ & 29.3$_{\downarrow11.2}$ & 34.3 \\
\midrule

ABDD
& 85.5 & \textbf{78.3}$_{\downarrow7.2}$ & \textbf{75.8}$_{\downarrow9.7}$ & \textbf{79.9}
& 86.3 & \textbf{81.1}$_{\downarrow5.2}$ & \textbf{79.7}$_{\downarrow6.6}$ & 82.4
& 47.4 & \textbf{43.9}$_{\downarrow3.5}$ & \textbf{42.0}$_{\downarrow5.4}$ & \textbf{44.4} \\
\bottomrule
\end{tabular}}
\vspace{-0.2cm}
\end{table}

\subsubsection{Qualitative Results}
Fig. 3 compares the detection results of the baseline and ABDD on target-domain images from CD-AeBD-I/II and HD-AeBD. (a) shows the ground-truth annotations. (b) presents the results of the baseline detector, which shares the same architecture as ABDD but has no domain adaptation strategy. (c) illustrates the results of ABDD. The baseline detector is sensitive to domain variations, including changes in illumination, viewpoint, and defect morphology, which leads to missed detections or misclassifications. The missed detections in rows 1, 2, and 7 are mainly caused by illumination variations. The missed detections of breakdown defects in rows 3 and 4 are associated with viewpoint changes. The misclassifications in rows 5 and 6 are likely caused by defect-morphology variations. In contrast, ABDD effectively adapts to the target domain at test time and produces more accurate detections. In addition, feature map visualization shows that ABDD effectively suppresses defect-irrelevant features in shallow feature maps. This suppression enables subsequent deep layers to focus on and enhance regions of interest. This property is critical for detecting tiny defects such as breakdowns in rows 3–4 and defects in complex backgrounds such as scratches in row 7.

\begin{figure}
\centering
\includegraphics[width=0.95\textwidth]{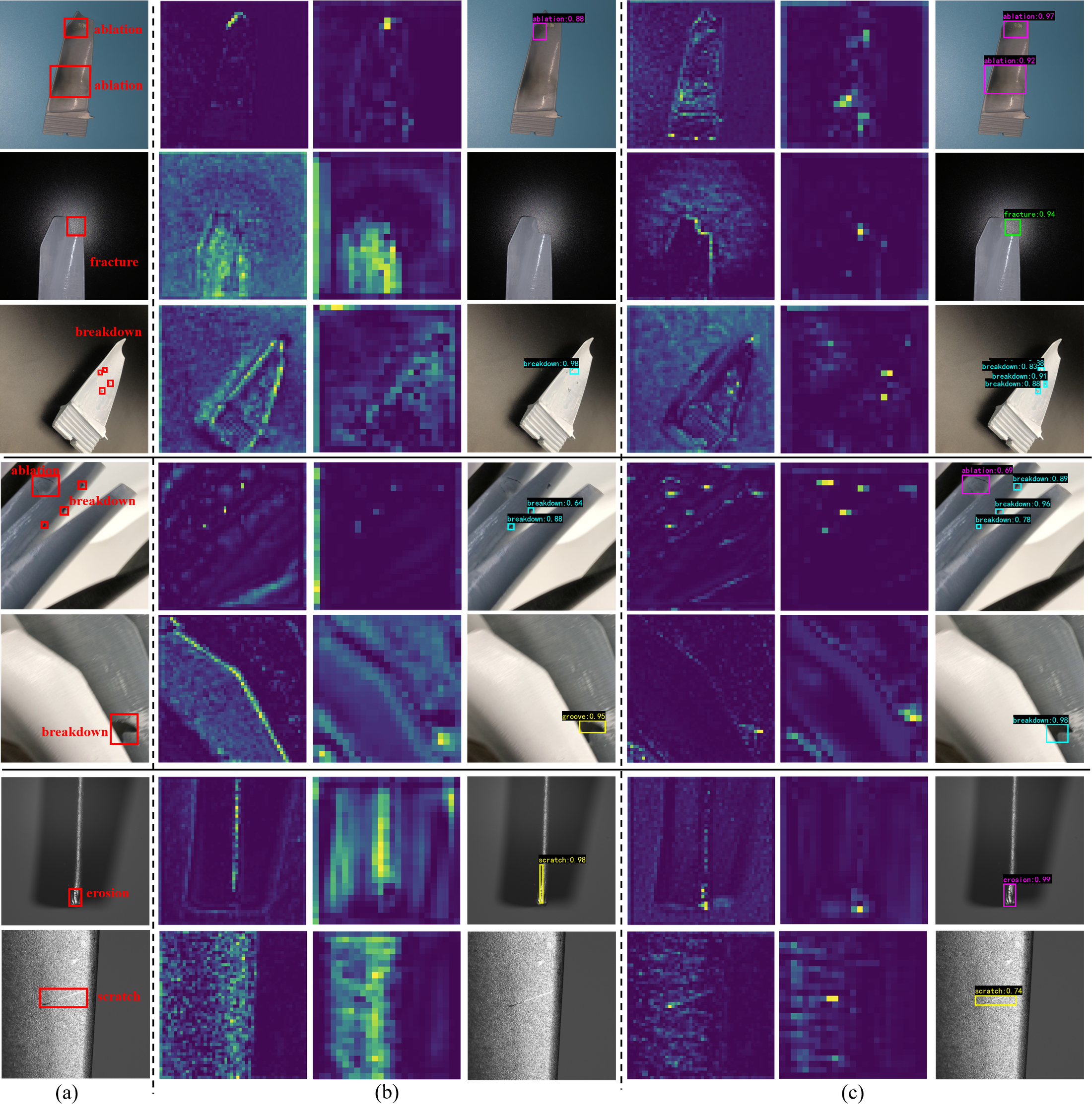}
\vspace{-0.15cm}
\caption{Visualizations of shallow/deep feature maps and detection results.  Rows 1--3 are target domains in CD-AeBD-I, rows 4--5 are target domains in CD-AeBD-II, and rows 6--7 are target domains in HD-AeBD.}
\end{figure}

\begin{table}
\centering
\caption{Ablation study of ABDD. CD-I, CD-II, and HD denote average performance drops, where lower is better. Aux. reports GT$_p$ for UBF and trainable parameters for tuning. Best results are in \textbf{bold}.}
\label{tab_ablation}
\scriptsize
\setlength{\tabcolsep}{3.2pt}
\renewcommand{\arraystretch}{0.86}
\begin{tabular}{@{}llcccl@{}}
\toprule
Comp. & Variant & CD-I & CD-II & HD & Aux. \\
\midrule
DAS & w/o PB, FD & 13.1 & 25.7 & 10.1 & -- \\
DAS & PB only & 6.9 & 24.4 & 6.5 & -- \\
DAS & FD only & 7.4 & 18.9 & 11.2 & -- \\
DAS & PB + FD & \textbf{5.1} & \textbf{8.5} & \textbf{5.9} & -- \\
\midrule
UBF & Conf. & 10.9 & 38.3 & 12.5 & 35.4/18.8/33.9 \\
UBF & PE + CE & 6.2 & 17.6 & 8.8 & 60.2/39.7/41.2 \\
UBF & Conf. + PE & 6.4 & 14.0 & 7.9 & 63.6/45.0/77.0 \\
UBF & Conf. + PE + CE & \textbf{5.1} & \textbf{8.5} & \textbf{5.9} & \textbf{64.8}/\textbf{48.1}/\textbf{77.2} \\
\midrule
Tune & Full & 14.3 & 30.5 & 10.8 & 27.7M/100\% \\
Tune & LayerNorm & 13.7 & 26.1 & 14.1 & 0.03M/0.10\% \\
Tune & Adapter & 11.1 & 16.0 & 10.2 & 0.15M/0.50\% \\
Tune & Mona & 5.4 & \textbf{8.3} & 7.8 & 1.38M/4.98\% \\
Tune & SDM & \textbf{5.1} & 8.5 & \textbf{5.9} & 0.64M/2.31\% \\
\bottomrule
\end{tabular}
\end{table}

\subsection{Ablation Study}
Table 4 summarizes the ablation results of the three key components in ABDD, including the dual-alignment strategy, uncertainty-aware pseudo-box filtering, and parameter-efficient tuning module. In the Aux. column, GT$_p$ denotes the precision of retained pseudo boxes on CD-I/CD-II/HD for offline pseudo-label quality analysis, and the percentages in the tuning variants denote the ratio of trainable parameters relative to full fine-tuning. The results show that using either pseudo-box alignment or feature-distribution alignment alone can reduce domain-shift degradation, but their combination achieves the most stable performance across CD-AeBD I, CD-AeBD II, and HD-AeBD. For pseudo-box filtering, confidence-only filtering produces many poorly localized pseudo boxes, whereas adding positional entropy and classification entropy improves pseudo-box quality and further reduces performance degradation. For the tuning module, full fine-tuning and LayerNorm updating are less effective, suggesting that excessive or insufficient parameter updates may harm adaptation. SDM achieves the best overall trade-off, obtaining the lowest degradation on CD-AeBD I and HD-AeBD while using only 0.64M trainable parameters, which is 46.4\% of Mona. Additional experimental results and hyperparameter sensitivity analyses are provided in \textit{Supplementary Material~IV}.

\section{Discussion and Conclusion}
This paper proposes ABDD, a high-precision Aero-engine Blade Defect Detector for robust test-time adaptation in automated aero-engine blade inspection. To mitigate performance degradation caused by domain shifts such as production batches, imaging views, and illumination variations, ABDD combines a Dual-Alignment Strategy (DAS) for feature-statistic and pseudo-box alignment, Uncertainty-aware Box Filtering (UBF) for reliable pseudo-label selection, and Sparse Dilated Mona (SDM) for parameter-efficient transformer adaptation. Experiments on CD-AeBD and HD-AeBD demonstrate that ABDD effectively reduces domain-shift degradation, achieving average $mAP@50$ improvements of 5.6\%, 11.5\%, and 2.8\% on CD-AeBD Subset I, CD-AeBD Subset II, and HD-AeBD, respectively, together with a 6.4\% gain in $mAP@50:95$ on HD-AeBD. The deployed inspection system further validates its practical value for rigorous blade screening and flight-safety-related manufacturing inspection. Nevertheless, ABDD still requires two forward propagations and one backward propagation during adaptation, which may limit its use in highly real-time scenarios. In addition, UBF mainly suppresses false-positive pseudo boxes but cannot explicitly recover false-negative defects under extremely large domain gaps. Future work will explore faster adaptation strategies and vision-language guidance with textual prompts to improve false-negative discovery and extend ABDD to broader industrial inspection scenarios.

\section*{Acknowledgements}
This work was supported in part by the National Natural Science Foundation of China No. 62473127, the S\&T Program of Hebei Province No. 242Q4302Z, the National Key Research and Development Program of China No. 2024YFB3310904, the Interdisciplinary Postgraduate Training Program of Hebei University of Technology No.HEBUT-Y-XKJC-2023002, and the China Scholarship Council No. 202506700027.
\bibliographystyle{splncs04}
\bibliography{references}

\end{document}